\documentclass[letterpaper, 10 pt, conference]{ieeeconf}  
\usepackage{cite}
\usepackage{amsmath,amssymb,amsfonts}
\usepackage{graphicx}
\usepackage{threeparttable}
\usepackage{booktabs}
\usepackage{makecell} 
\usepackage{multirow} 
\usepackage{textcomp}
\usepackage[colorlinks,linkcolor=blue,citecolor=black]{hyperref}
\usepackage{algpseudocode}
\usepackage{algorithm}
\usepackage{xcolor}
\usepackage{amsmath}
\usepackage{placeins}

\IEEEoverridecommandlockouts                              

\title{\LARGE \bf
MagicSkin: Balancing Marker and Markerless Modes in Vision-Based Tactile Sensors with a Translucent Skin
}

\author{Oluwatimilehin Tijani$^{1*}$, Zhuo Chen$^{1*}$, Jiankang Deng$^{2}$ and Shan Luo$^{1}$
\thanks{*This work was supported by the EPSRC project ``ViTac: Visual-Tactile Synergy for Handling Flexible Materials" (EP/T033517/2).}
\thanks{$^{1}$Oluwatimilehin Tijani, Zhuo Chen, and Shan Luo are with the Robot Perception Lab, Centre for Robotics Research, Department of Engineering, King's College London, London WC2R 2LS, United Kingdom. Emails: {\tt\small \{zhuo.7.chen ,shan.luo\}@kcl.ac.uk}.}
\thanks{$^{2}$Jianjang Deng is with Imperial College London, London SW7 2AZ, United Kingdom.}
\thanks{$*$Contribute equally to this work.}
}

\begin{document}
\maketitle
\thispagestyle{empty}
\pagestyle{empty}

\begin{abstract}

Vision-based tactile sensors (VBTS) face a fundamental trade-off in marker and markerless design on the tactile skin: opaque ink markers enable measurement of force and tangential displacement but completely occlude geometric features necessary for object and texture classification, while markerless skin preserves surface details but struggles in measuring tangential displacements effectively. Current practice to solve the above problem via UV lighting or virtual transfer using learning-based models introduces hardware complexity or computing burdens. This paper introduces MagicSkin, a novel tactile skin with translucent, tinted markers balancing the modes of marker and markerless for VBTS.  It enables simultaneous tangential displacement tracking, force prediction, and surface detail preservation. This skin is easy to plug into GelSight-family sensors without requiring additional hardware or software tools. We comprehensively evaluate MagicSkin in downstream tasks. The translucent markers impressively enhance rather than degrade sensing performance compared with traditional markerless and inked marker design: it achieves best performance in object classification (99.17\%), texture classification (93.51\%), tangential displacement tracking (97\% point retention) and force prediction (66\% improvement in total force error). These experimental results demonstrate that translucent skin eliminates the traditional performance trade-off in marker or markerless modes, paving the way for multimodal tactile sensing essential in tactile robotics. See videos at this \href{https://zhuochenn.github.io/MagicSkin_project/}{link}.

\end{abstract}

\section{INTRODUCTION}\label{intro}

Robotic manipulation requires rich tactile feedback to achieve human-level dexterity in tasks such as in-hand reorientation, slip detection, and adaptive grasping. Conventional force sensors offer limited spatial resolution and are prone to electromagnetic interference, restricting their industrial use. Vision-Based Tactile Sensors (VBTS), such as GelSight \cite{Yuan2017}, overcome these limitations by optically capturing soft elastomer deformation, allowing simultaneous measurement of forces, surface geometry, and material properties with high resolution and scalable fabrication.

\begin{figure}[htbp]
\centering
\includegraphics[width=0.43\textwidth]{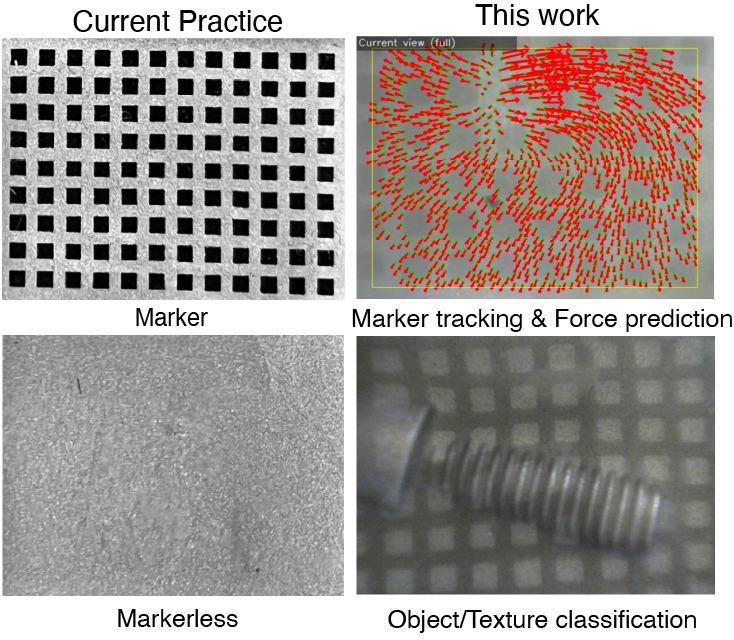}
\caption{Demonstration of MagicSkin. Current practices use dense ink design and clear skin for marker and markerless modes in VBTS respectively. Our Magic skin can balance above two modes in one translucent skin for marker tracking, force prediction and object/texture  classification without introducing extra hardware or software to switch.}
\label{fig_magicskin}
\end{figure}

 The traditional clear elastomer lacks features for tangential displacement tracking, which limits shear force measurement. Attempts to address this include the use of opaque ink markers \cite{Shimonomura2019}, micro-textures \cite{Vitrani2025, Fan2024}, and UV-sensitive markers \cite{Kim2022}. However, each introduces trade-offs: opaque markers obscure geometry and degrade texture classification \cite{Fang2024, Schmitz2011}; UV systems require added hardware; multicamera designs \cite{Cui2021, Padmanabha2020, Zhang2022} increase complexity; and digital transfer \cite{Ou2024} increases computational burdens. It is still challenging to balance the marker and markerless modes in VBTS to achieve accurate classification, measurement of tangential displacement, and force with high performance while avoiding complexity.

This work introduces MagicSkin, a novel tactile skin that integrates translucent, tinted markers into the elastomer surface to address the limitations of existing marker-based and markerless VBTS designs. As shown in Fig.~\ref{fig_magicskin}, this approach provides visible marker squares for tangential displacement and force estimation while preserving fine-grained surface details for classification. The key contributions are as follows:

\begin{enumerate}
    \item We propose a translucent marker architecture that eliminates the trade-off between marker-based and markerless sensing in VBTS, requiring no additional hardware or software modifications. This enables simultaneous recovery of fine object geometry and robust displacement features.  
    \item We validate the design on object and texture classification, achieving up to 99.2\% accuracy and 93.5\% texture recognition, outperforming both dense-ink markers (98.3\% / 63.9\%) and clear elastomer markerless sensors (98.4\% / 85.0\%).  
    \item We demonstrate superior tangential tracking performance, with a 54\% increase in retention and only 0.014~px forward–backward error compared to dense ink markers.  
    \item We show improved 3-axis force prediction, reducing mean absolute error by over 66\% relative to dense ink markers with sequential images, enabling accurate estimation of both normal and shear forces.  
\end{enumerate}

The rest of the paper is structured as follows: 
Section~\ref{Sec:relatedworks} provides an overview of related works; Section~\ref{Sec:design&fabrication} details the design and fabrication of translucent skin ; Section~\ref{Sec:tactile_image_processing} introduce tactile image processing; Section~\ref{Sec:data_collect} details our data collection process for three downstream tasks; Section~\ref{sec:experiment} analyses the experimental results. Finally, Section~\ref{Sec:DiscussionConclusion} summarises this work.

\section{RELATED WORK}\label{Sec:relatedworks}

\subsection{Vision-Based Tactile Sensors}\label{R1}

Vision-based tactile sensors (VBTS) integrate a deformable elastomeric membrane, an embedded camera, and a controlled illumination system to convert physical contact into rich visual signals. When the elastomer is indented by an external object, its surface deforms and interacts with the lighting to produce variations in shading, texture, and reflection. These optical cues are recorded by the internal camera and subsequently processed for perception tasks. For instance, GelFlex employs such a configuration to achieve object classification when mounted on a robotic finger \cite{she2020}. The GelSight family of sensors has demonstrated applications ranging from texture recognition with ink-marked membranes \cite{cao2020} to high-fidelity three-dimensional surface reconstruction \cite{Lu2025}.

\subsection{Marker-Based Force Estimation}\label{R3}

Force estimation requires engineered visual references \cite{Li2023} in VBTS. Traditional opaque dense ink markers \cite{Shimonomura2019, Yuan2015} are designed underneath the elastomer to enhance measurement of tangential displacement, but they mask local deformations, impairing texture and material recognition \cite{Xin2025}. 3D embedded markers \cite{Fan2025} integrate cubic microgrids via 3D printing, enabling rich force estimation, but computational overhead limits real-time operation. Recent shadow-based marker design \cite{Vitrani2025} employs micrometer-scale dimples whose shadows encode tangential displacements. However, these cues collapse under high compression, restricting their operational range.   

\subsection{Strategies to Combine Marker and Markerless Modes}\label{R4}

To address the occlusion of surface detail by markers, several strategies have been proposed:  

\noindent
\textbf{UV-Reactive Markers.} Fluorescent patterns revealed under UV illumination \cite{Wang2022, Abad2020, Kim2022} enable toggling between geometry and tracking. However, this requires additional hardware and prevents the simultaneous capture of both modalities.  

\noindent
\textbf{Multiple Elastomer Layers.} ChromaTouch \cite{Lin2020} uses stacked coloured elastomers, where the hue encodes the vertical distance and the centroid shifts encode the shear displacement. While effective, the colour-based scheme conflicts with multi-colour illumination used in photometric stereo.  

\noindent
\textbf{Digital Transfer.} Learning-based dual-modality transfer method\cite{Ou2024} can translate the tactile skin across marker or markerless modes. However, this approach is highly data-driven and increases computational burdens. The marker-to-markerless mode discards fine deformation details beneath the markers after inpainting.  

\subsection{Current Limitations and Research Gap}\label{R5}

Despite diverse innovations, existing approaches still impose compromises between robust tangential tracking, force prediction and faithful recovery of surface geometry. Opaque markers maximize trackability but occlude deformation detail; shadow- and light-based features suffer under compression or generalization shifts; and hybrid solutions often require complex hardware or lose information. As summarized in Table~\ref{tab:vbts_comparison}, current designs do not simultaneously preserve surface detail and enable reliable tangential tracking. This motivates the development of new elastomer–marker design that sustain, geometric fidelity, robust marker displacement track, and high-accuracy force prediction.  

\begin{table*}[h!]
\centering
\caption{Comparison of Vision-Based Tactile Sensor Designs Across Key Tasks}
\label{tab:vbts_comparison}
\resizebox{\textwidth}{!}{%
\begin{tabular}{|c|c|c|c|c|c|c|c|}
\hline
\textbf{Design (with References)} & \textbf{Object / Texture Classification} & \textbf{Tangential Tracking} & \textbf{Force Estimation} & \textbf{Cost} & \textbf{Complexity} & \textbf{Mode} & \textbf{Mode Switching} \\ \hline
Ink Markers \cite{Yuan2015}        & Limited (occluded detail)   & Yes, robust          & Good for shear  & Low   & Low    & Single  & N/A \\ \hline
3D Embedded Markers \cite{Fan2025} & Medium (partial occlusion)  & Yes, high compute cost& Good (0-2N)                  & Medium& High   & Single  & N/A \\ \hline
Shadow Dimples \cite{Vitrani2025}     & Medium                      & Yes, limited range  & Good (0-2N)                  & Medium& Medium & Single  & N/A \\ \hline
UV Markers \cite{Wang2022, Abad2020, Kim2022}         & High (markerless mode)      & Yes, high compute cost        & Good (high compute cost)         & High & High & Dual   & Requires UV hardware \\ \hline
Virtual Transfer \cite{Ou2024}      & High (markerless mode)       & Yes, unstable   & unknown & Medium & Medium & Dual & Learning-based model \\ \hline
\textbf{This Work} & \textbf{High (99.2\% accuracy)} & \textbf{High (54\% retention gain, 0.014 PX error)} & \textbf{Good (0-6N, 66\% lower error)} & \textbf{Low} & \textbf{Low} & \textbf{Dual} & \textbf{no extra hardware/software} \\ \hline
\end{tabular}%
}
\end{table*}


\section{Design and Fabrication}
\label{Sec:design&fabrication}
\subsection{Design Rationale}\label{Design Rationale}

The tactile skin design prioritizes optical clarity, marker visibility, and displacement tracking accuracy, addressing the challenge of achieving sufficient contrast while preserving transparency and mechanical compliance. Two configurations as shown in Fig. \ref{fig_fabrication}B were fabricated to study marker-markerless trade-offs: grey squares with clear lines and clear squares with grey borders, both implemented as 9×12 grids of 1.0 mm squares with 1.0 mm spacing. The grey-square design leaves most of the surface clear, while the grey-line design tints most of the surface, creating a controlled variation in tinted coverage that supports direct testing of potential trade-offs between surface detail preservation and lateral displacement tracking. Positioning markers directly beneath the pigment layer on a single 2.0 mm backing (rather than between dual 1.0 mm backings) reduced light scattering and enhanced RGB contrast for improved tracking accuracy.

\subsection{Fabrication of Elastomer}\label{Fabrication}

\begin{figure}[htbp]
\centering
\includegraphics[width=0.35\textwidth]{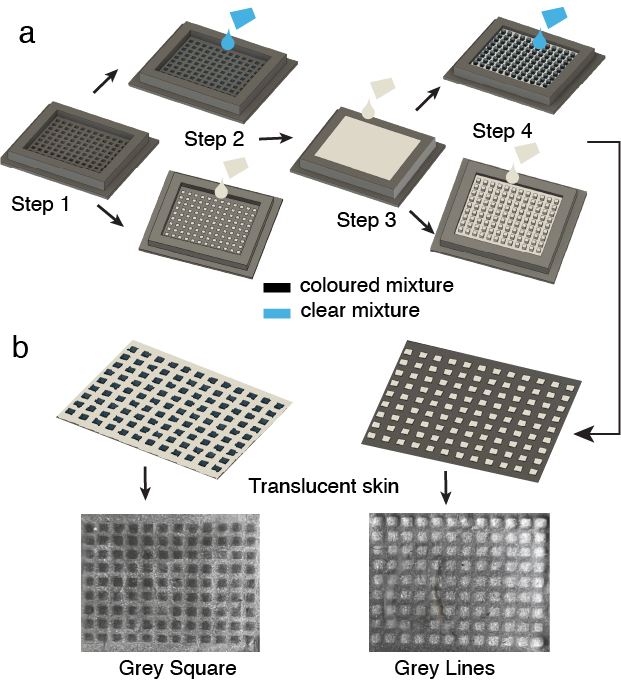}
\caption{Fabrication process (a) and two designed translucent skins (b).}
\label{fig_fabrication}
\end{figure}

Elastomers were fabricated using a staged casting process to produce the two grid configurations shown in Fig.~\ref{fig_fabrication}. All variants employed XP-565 silicone mixed at a ratio of 1 g Part A to 12 g Part B. The acrylic mould assembly comprised a 1.0~mm grid guide, a 3.0~mm thickness former, and a smooth base plate to ensure uniform surfaces.  

\subsubsection{Base Silicone Preparation}\label{Fabrication:1}

\textbf{Step 1:} The mould assembly was prepared with the grid guide, thickness former, and base plate, establishing the framework for subsequent casting.  

\subsubsection{Clear Squares with Grey Lines}\label{Fabrication:2}

\textbf{Step 2:} Clear silicone was poured into the grid cutouts, filled to the mould edge, and degassed to release trapped air, with additional mixture replenishing the volume lost during bubble escape. The mould was then cured at 70\(^\circ\)C for 30 minutes inside a UV-shielded enclosure.  
\textbf{Step 3:} After curing, the grid guide was removed and the gaps between the clear squares were filled with tinted silicone, prepared by lightly doping the base mixture with Silc Pig black ink. Degassing and curing at 70\(^\circ\)C for 30 minutes were repeated to complete the design.  
\textbf{Step 4:} This step is not required for the clear-square configuration, as the tinted regions are already formed by backfilling the gaps left after Step 3.  

\subsubsection{Grey Squares with Clear Lines}\label{Fabrication:3}

\textbf{Step 2:} Pigmented silicone (Silc Pig black ink) was poured into the square cutouts, degassed, and cured at 70\(^\circ\)C for 30 minutes.  
\textbf{Step 3:} Clear silicone was added to fill the remaining mould space, followed by degassing and a second curing cycle at 70\(^\circ\)C for 30 minutes.  
\textbf{Step 4:} The elastomer was inverted so that the tinted squares faced upward, and a final clear layer was applied and smoothed flush between the squares. Degassing and curing at 70\(^\circ\)C for 30 minutes were repeated to achieve a uniform surface.  

\subsubsection{Opaque Backing Layer}\label{Fabrication:4}

\textbf{Step 5:} To enhance optical contrast and suppress background light, an opaque backing was applied to all elastomer variants. The mixture was prepared from XP-565 silicone (1 g Part A : 12 g Part B) combined with 4.0 g aluminium powder (1~µm particle size) and 4.0 g Silver Cast Magic pigment for reflectivity control. To adjust viscosity, 20.0 g of Novocs solvent was added in two 10 g increments during mixing. A thin uniform layer of this compound was poured over the cured elastomer and subjected to degassing followed by curing at 70\(^\circ\)C for 30 minutes. The resulting layer provided optical opacity while maintaining mechanical compliance and sensitivity to fine deformations.  

\subsubsection{Final Assembly}\label{Fabrication:5}
The completed elastomer exhibits 3.0 mm total thickness comprising an opaque pigment layer for light reflection, a marker layer with the specific grid pattern, and clear backing layers. This configuration positions markers immediately adjacent to the contact layer, maximizing deformation sensitivity while protecting the grid from direct wear through repeated contact cycles.

\subsection{Sensor Assembly}\label{Illumination, Camera, and Assembly}

\begin{figure}[htbp]
\centering
\includegraphics[width=0.43\textwidth]{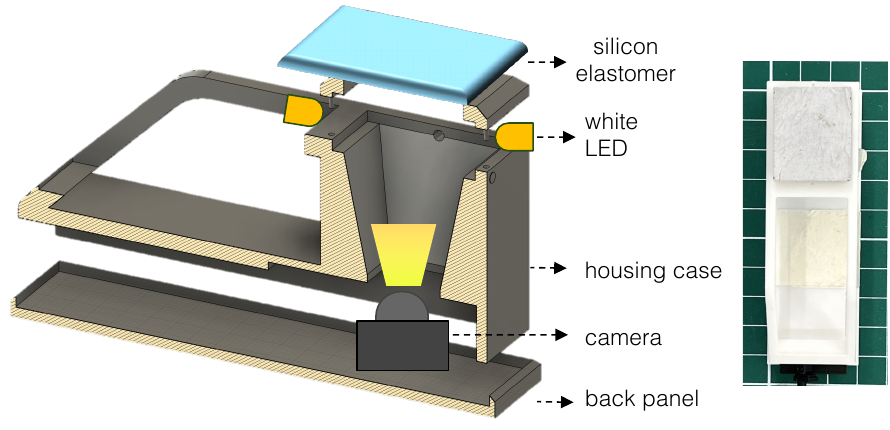}
\caption{Sensor assembly.}
\label{fig_sensor}
\end{figure}

As shown in Fig.~\ref{fig_sensor}, the sensor assembly integrates the elastomer with a protective housing to ensure stable illumination and consistent optical performance. The elastomer is mounted on a 3.0~mm clear acrylic plate and recessed 2.0~mm into a white PLA casing (Bambu Lab P1S), which provides structural stability, uniform stress distribution, and reproducible contact conditions. An RGB camera is positioned 24.0~mm from the contact surface to maximize coverage within the depth of field, while six white LEDs are arranged in pairs on three sides of the housing to deliver uniform illumination. Videos were recorded at 640×480 resolution, supporting real-time processing and enabling sub-pixel marker tracking.

\section{Tactile Image Processing}\label{Sec:tactile_image_processing}

The image processing for tangential displacement tracking involves preprocessing each frame capturing the elastomer's surface with the markers visible before measuring any displacement that may have occurred from the initial frame. This Lucas-Kanade optical flow tracking experiment is designed to robustly track key-points on an elastomer design in real-time camera footage using a preprocessing pipeline that includes 5\% border cropping, gray-world white balance correction, per-channel Retinex normalization, and enhancement techniques like Contrast Limited Adaptive Histogram Equalization (CLAHE) \cite{khairi2025} and unsharp masking. The system employs a two-stage masking strategy with a primary geometry-aware mask optimized for square patterns, validated through health checks for mask quality. Feature tracking is performed using GFTT key-point detection (up to 1000 points) \cite{Shi1994} combined with pyramidal Lucas-Kanade optical flow (155×155 window size, 3 pyramid levels) and forward-backward error estimation to assess tracking quality, while continuously monitoring performance metrics including retention rate (percentage of successfully tracked key-points) and forward-backward error (average pixel displacement when tracking backwards).

\section{Data Collection}\label{Sec:data_collect}

\subsection{Classification Task}\label{Cls_data}

Images were acquired at 640×480 resolution under standardized white LED illumination with the camera positioned 24.0 mm from the contact surface. Captured images that include significant motion blur are excluded.

\begin{figure}[htbp]
\centering
\includegraphics[width=0.40\textwidth]{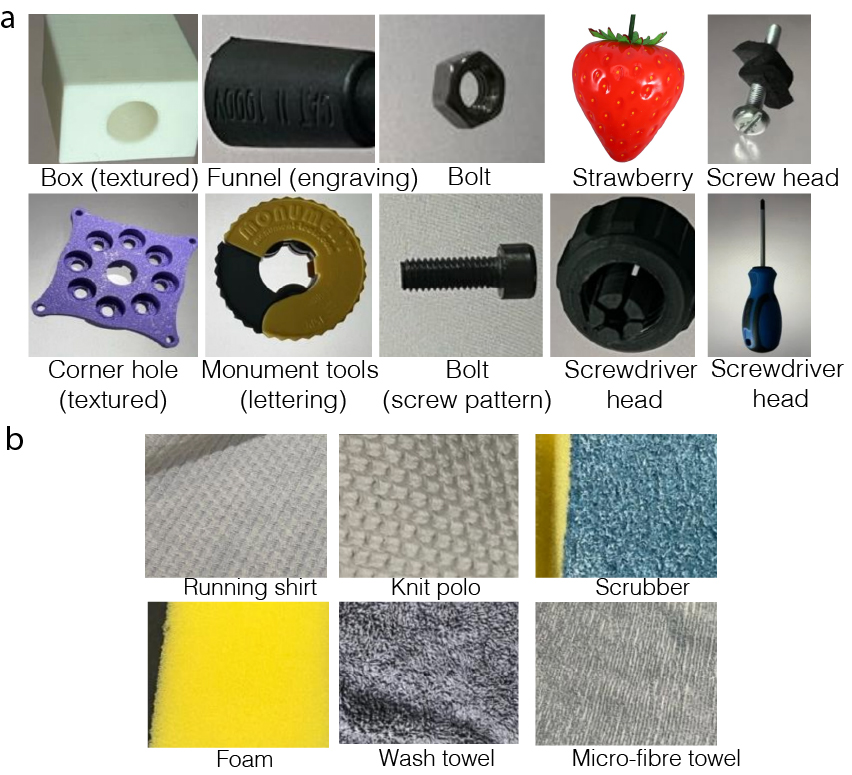}
\caption{Objects used in object classification (a) and texture recognition (b)}
\label{fig_10obeject}
\end{figure}

\begin{figure}[htbp]
\centering
\includegraphics[width=0.48\textwidth]{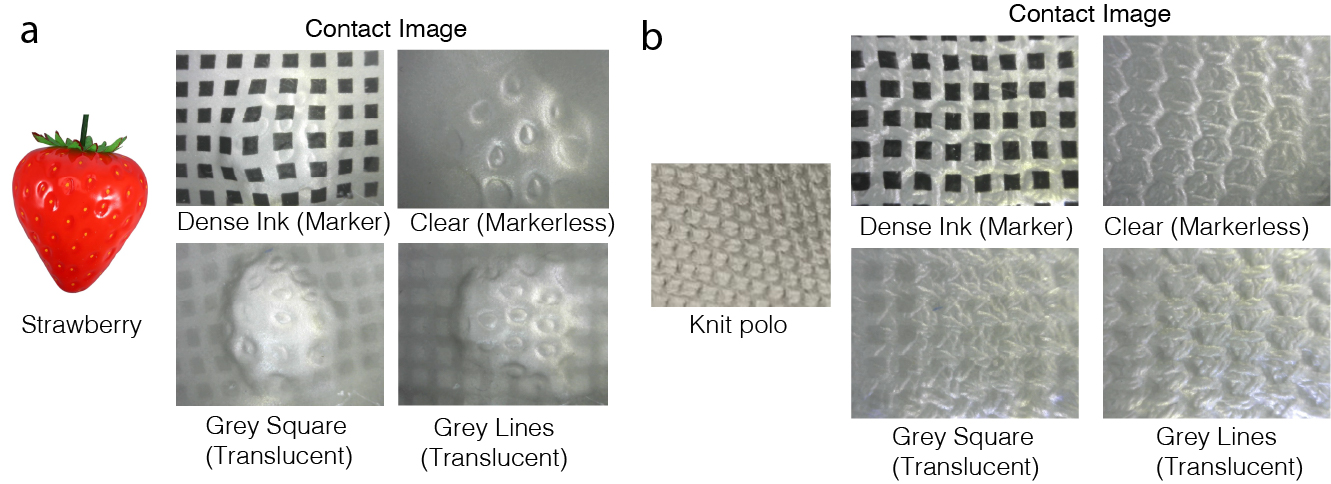}
\caption{Comparison of tactile image for object classification (a) and texture recognition (b) using different tactile skin design.}
\label{fig_obeject}
\end{figure}

\subsubsection{Object Classification}\label{object_classification_data_collection}
Object classification employed ten diverse objects (box, funnel, bolt, strawberry, screw head, corner hole plate, monument tools) representing varied geometric complexity and surface properties as shown in Fig. \ref{fig_10obeject}a. Each object was systematically contacted at nine locations distributed across a 3×3 grid partitioning of the elastomer surface. Images were captured at 0.1-second intervals from maximum applied force until contact cessation, generating 2,000 images per object per sensor configuration (80,000 total images across four elastomer designs). Examples of tactile images contacting with a strawberry are shown in Fig. \ref{fig_obeject}a.

\subsubsection{Texture Classification}\label{texture_classification_data_collection}
Texture classification employed six categories (foam, knit polo, microfibre cloth, running shirt, scrubber, wash towel) selected to represent diverse tactile discrimination challenges as shown in Fig. \ref{fig_10obeject}b. Data collection followed a two-phase protocol: Phase I captured ground-truth texture responses through controlled flat-sheet contact where a flat acrylic slate was placed on top of texture materials and force was applied until the texture pattern was most visible, then gradually reduced until contact cessation. Contact was applied across a 3×3 grid positioning system. Phase II employed texture samples wrapped around cylindrical (25 mm diameter) and spherical (30 mm diameter) forms to simulate realistic three-dimensional manipulation scenarios. The dataset comprises 18,000 images with 3,000 images per texture category distributed equally across four sensor configurations. Consistent environmental conditions were maintained throughout data acquisition, and quality control procedures eliminated images with insufficient contact pressure. Examples of tactile images contacting with Knit polo are shown in Fig. \ref{fig_obeject}b.

\subsection{Optical Flow Tracking}\label{optical_flow_tracking}

Data were collected using a 3×3 grid positioning system across the elastomer surface, with five motion patterns executed at each grid cell: horizontal, vertical, and diagonal displacements with alternating directions, normal loading at the centre, and circular motion originating from the centre. Videos were recorded with an RGB camera 24.0~mm from the contact surface at 640×480 resolution under white LED illumination. Contact scenarios included varying force levels (light, moderate, heavy), single versus multiple contact points, and systematic deformation patterns to evaluate tracking robustness.  

\subsection{Force Prediction}\label{Force_data}

\subsubsection{Setup}\label{setup_force}

\begin{figure}[htbp]
\centering
\includegraphics[width=0.45\textwidth]{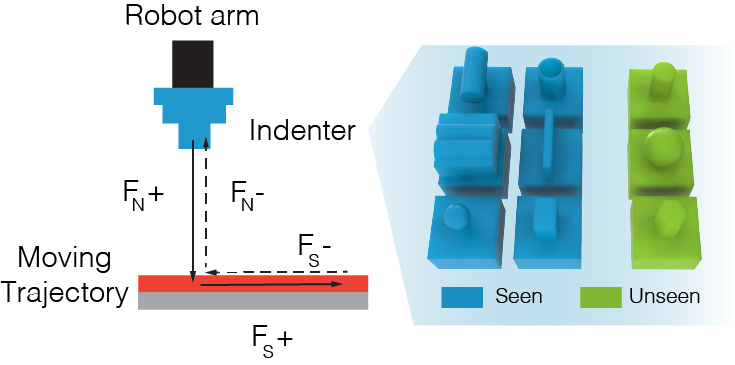}
\caption{Data collection process for force prediction. A robot arm equipped with indenters (seen and unseen) moves in four steps with vertical and tangential movement with different forces.}
\label{fig_indenter}
\end{figure}

The data collection setup consists of four main components: a robot arm, an ATI Nano17 force/torque sensor, 3D-printed indenters and tactile sensors. The indenters comprise 6 indenters in the seen group and 3 in the unseen group, as shown in Fig. \ref{fig_indenter}. Each indenter was mounted onto the robot arm with a fixed initial position and orientation. 

\subsubsection{Trajectory}\label{setup_force}

We uniformly selected five contact locations in the middle and edges to provide comprehensive coverage of the soft elastomer surface. At each location, the indenter executed a four-phase motion toward nearby targets in the normal and shear directions as shown in Fig. \ref{fig_indenter}: (i) moving downward, (ii) moving tangentially, (iii) tangential return, and (iv) moving upward. This captures both increasing and decreasing force trajectories \((F_N^+, F_S^+)\) and \((F_N^-, F_S^-)\), where \(N\) and \(S\) denote normal force and shear force components, respectively. The motion speed was set to \(20\%\) in the UR5e panel. The images and forces were synchronized and acquired at \(40\,\text{Hz}\) in real time across all trials. 

\subsubsection{Parameters \& Data Amount}\label{setup_force}

For each contact location, the indenters applied steps of 0.3 mm to a maximum depth of 1.2 mm. In each step, the moving angle is $30^\circ$ to repeat 12 times in different directions with a shear distance of 1.0 mm. This configuration yielded 240 target points that span diverse directions and locations. In total, 60,000 tactile images were collected for four skins, respectively.

\section{EXPERIMENTAL RESULTS}\label{Sec:experiment}
\label{sec:experiment}

\subsection{Evaluation Metrics}

The evaluation framework employed three groups of metrics that were aligned with the core tasks of this study.  
(1) \textbf{Object and Texture Classification:} performance was quantified by classification accuracy using an EfficientNet-B2 backbone across ten objects (80,000 images) and six textures (18,000 images).  
(2) \textbf{Tangential Displacement Tracking:} precision was assessed using the forward–backward (FB) optical flow error in pixels and the percentage of point retention across sequences.  
(3) \textbf{Force Prediction:} normal and shear force estimation was evaluated against ground-truth measurements from an ATI Nano17 F/T sensor, using mean absolute error (MAE) and coefficient of determination ($R^2$).  

All experiments were conducted under standardized conditions with uniform white LED illumination, fixed camera placement at 24.0~mm, and a white PLA housing for light diffusion. Four sensor configurations were systematically compared: clear silicone (markerless, theoretical upper bound), dense black ink markers (baseline), and two variants of translucent markers (grey squares and grey lines, both implemented as 9×12 grids).  

\subsection{Object Classification Performance}\label{Object Classification Performance}

For each object–sensor pair, 1,200 images were collected and split into 70\% training, 30\% validation, and an independent test set of 800 images. The models were trained with EfficientNet-B2 over 50 epochs with early stopping (patience = 5), batch size 64, and repeated five times for statistical reliability.  

Table \ref{tab:effnetb2_sensor_design} summarizes the results. The grey squares design achieved the best performance, with 99.17\% mean accuracy ($\pm$0.13\%) and worst-case accuracy above 99.0\%, surpassing both the markerless (98.39\%) and dense ink (98.28\%) baselines. Structured translucent markers enhanced rather than degraded classification, with both line- and square-based designs outperforming the clear elastomer. Dense ink markers delivered the weakest and most variable results, while grey squares provided the most consistent accuracy across runs, confirming that structured markers improve discriminative feature learning and robustness.  

\begin{table}[h!]
\centering
\caption{Comparison of Sensor Design Performance using EfficientNet-B2 \\
Note: Accuracy and Macro-F1 are reported across all experimental runs}
\label{tab:effnetb2_sensor_design}
\resizebox{0.48\textwidth}{!}{%
\begin{tabular}{|c|c|c|c|c|c|}
\hline
\textbf{Sensor Design} & \textbf{Best Accuracy} & \textbf{Average Accuracy} & \textbf{Std. Dev.} & \textbf{Worst-Case Acc.} & \textbf{Avg Macro-F1} \\ \hline
Clear (markerless)     & 98.64\% & 98.39\% & $\pm$0.18\% & 98.21\% & 98.39\% \\ \hline
Dense Ink Markers      & 98.88\% & 98.28\% & $\pm$0.53\% & 97.75\% & 98.27\% \\ \hline
Grey Lines             & 99.21\% & 98.86\% & $\pm$0.36\% & 98.50\% & 98.86\% \\ \hline
Grey Squares           & \textbf{99.34\%} & \textbf{99.17\%} & \textbf{$\pm$0.13\%} & \textbf{99.04\%} & \textbf{99.17\%} \\ \hline
\end{tabular}%
}
\end{table}

\subsection{Texture Classification Performance}\label{Texture Classification Performance}

Each texture dataset was split into 70\% training, 30\% validation, and an independent test set of 250 images per class. Models were trained with EfficientNet-B2 for 50 epochs using early stopping (patience = 3) and batch size 128, with five independent runs per elastomer design for robustness.  

Table \ref{tab:texture_classification_comparison} shows that the grey lines design achieved the highest performance, with 93.51\% average accuracy and 93.65\% Macro-F1. This represents an +8.5 percentage point gain over the clear elastomer, +5.1 points over grey squares, and nearly +30 points over dense ink, which performed worst (63.87\% accuracy, 58.84\% F1). These findings indicate that sparse translucent markers preserve fine surface detail while providing reference features, enabling balanced and reliable classification across diverse textures.  

\begin{table}[h!]
\centering
\caption{Texture Classification Performance Across Sensor Designs \\
Note: Acc = Accuracy, F1 = Macro-F1 score}
\label{tab:texture_classification_comparison}
\resizebox{0.48\textwidth}{!}{%
\begin{tabular}{|c|cc|cc|cc|cc|}
\hline
\multirow{2}{*}{Metric} & \multicolumn{2}{c|}{Grey Lines} & \multicolumn{2}{c|}{Grey Squares} & \multicolumn{2}{c|}{Clear} & \multicolumn{2}{c|}{Dense Ink} \\ \cline{2-9} 
 & Acc & F1 & Acc & F1 & Acc & F1 & Acc & F1 \\ \hline
Best  & \textbf{0.9520} & \textbf{0.9528} & 0.9000 & 0.8901 & 0.9020 & 0.8994 & 0.6813 & 0.6326 \\ \hline
Avg   & \textbf{0.9351} & \textbf{0.9365} & 0.8842 & 0.8783 & 0.8500 & 0.8418 & 0.6387 & 0.5884 \\ \hline
\end{tabular}%
}
\end{table}

\subsection{Tangential Displacement Tracking Performance}\label{tangential Displacement Tracking Performance}

Table \ref{tab:sensor_performance} reports tangential displacement tracking across four marker-based sensor designs, evaluated using forward–backward (FB) error and point retention under varied contact directions, forces, and speeds.  

\begin{table}[h!]
\centering
\caption{Optical flow tracking performance across sensor designs. 
FB error = forward–backward optical flow error.}
\label{tab:sensor_performance}
\resizebox{0.48\textwidth}{!}{%
\begin{tabular}{|c|cccc|c|}
\hline
\textbf{Design} & \textbf{Mean} & \textbf{Std} & \textbf{Min} & \textbf{Max} & \textbf{Retention} \\ \hline
Dense Ink        & 0.017 & 0.033 & 0.000 & 0.258 & 62.9\%  \\ \hline
Grey Lines       & 0.020 & 0.026 & 0.001 & 0.170 & 76.4\%  \\ \hline
Grey Lines (Worn)& 0.038 & 0.052 & 0.002 & 0.442 & 86.3\%  \\ \hline
Grey Squares     & \textbf{0.014} & \textbf{0.020} & \textbf{0.002} & \textbf{0.118} & \textbf{97.0\%}  \\ \hline
\end{tabular}%
}
\end{table}

\begin{figure}[htbp]
\centering
\includegraphics[width=0.48\textwidth]{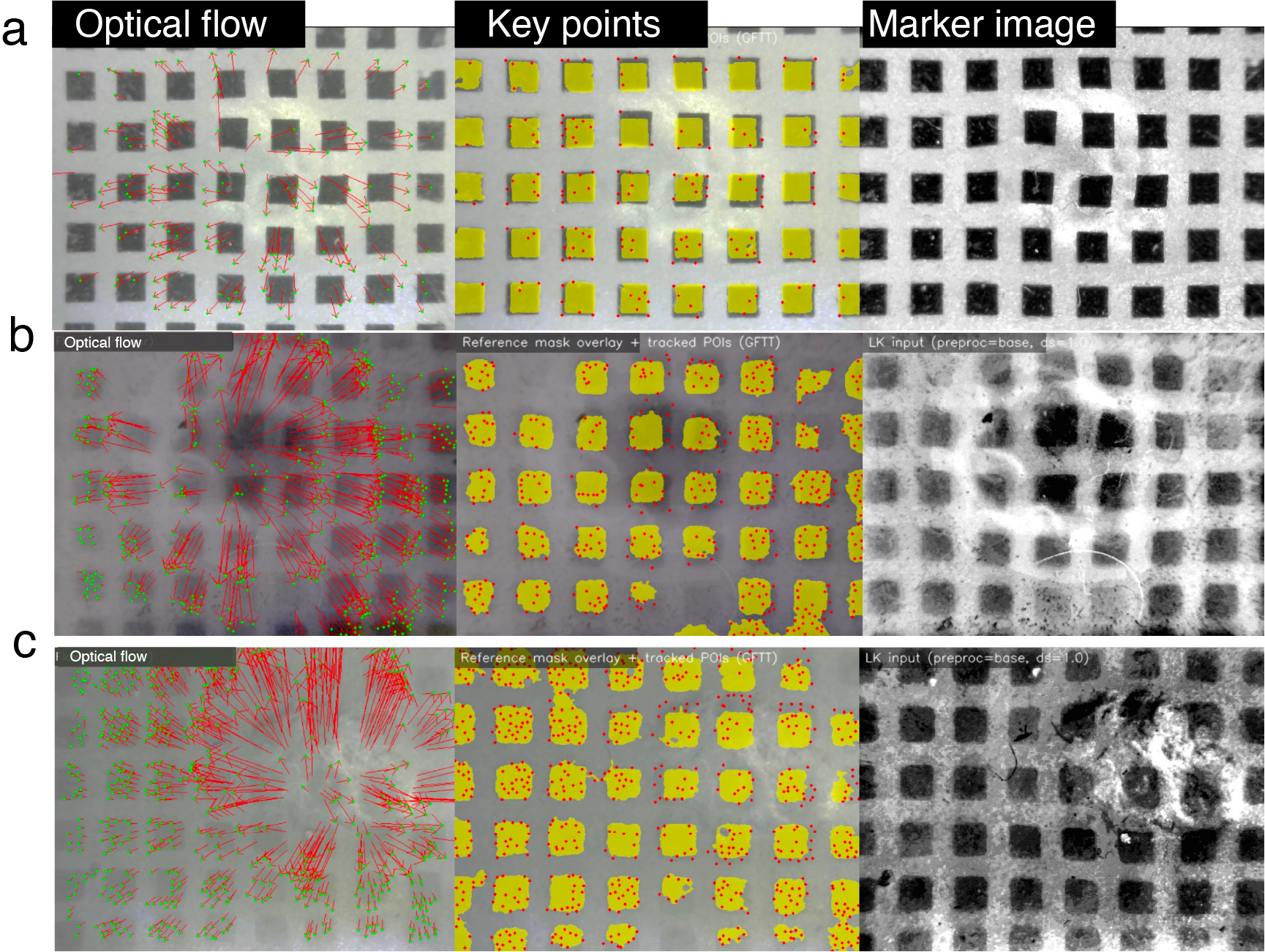}
\caption{Optical flow for measurement of tangential displacement by using dense ink markers (a), grey lines (b) and grey squares (c) respectively.}
\label{fig_optical_flow}
\end{figure}


The grey squares configuration delivered the most reliable performance, with the lowest mean FB error (0.014 PX), lowest variance (0.020 PX), and highest retention (97.0\%), confirming both accuracy and robustness (Fig.~\ref{fig_optical_flow}). Dense ink achieved low error in zero-contact conditions (0.000 PX minimum) but deteriorated under deformation, with mean error rising to 0.017 PX and retention dropping to 62.9\%. Grey lines offered moderate results (0.020 PX mean error, 76.4\% retention), though wear substantially degraded precision (0.038 PX mean error, 0.052 PX variance) despite a nominal rise in retention.  

Overall, translucent markers—and especially grey squares—proved the most practical and robust solution for displacement tracking, while dense ink was fragile under deformation and line markers degraded significantly with wear.  

\subsection{Force Prediction Performance}\label{Force_Performance}

To evaluate the force prediction performance by using both a single image and sequential images, we use ResNet and ResNet+ConvGRU \cite{chen2025general}. Both models are trained with seen indenters for 20 epochs with a learning rate of 0.1 and for another 20 epochs of a learning rate 0.001.

\FloatBarrier 
\begin{figure*}[h!]
\centering
\includegraphics[width=0.9\textwidth,height=0.25\textheight,keepaspectratio]{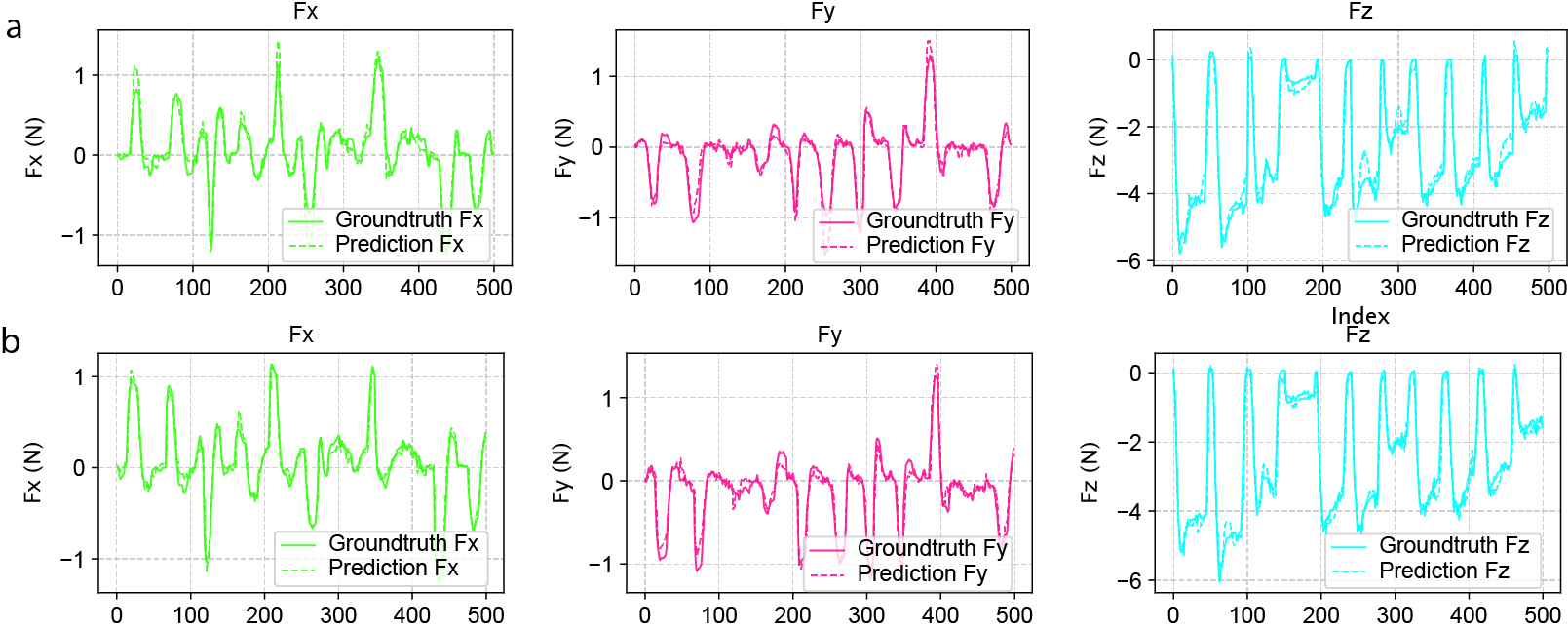}
\caption{Force prediction performance of grid line (a) and grid square (b) in seen objects (500 images). Ground truth is from ATI nano 17 F/T sensor.}
\label{fig_force_500}
\end{figure*}

Table. \ref{tab:seen} and Table. \ref{tab:unseen} illustrates the force prediction performance for four skins test in $seen$ indenters and $unseen$ indenters, respectively. It is impressively to find that translucent design with Grey squares demonstrates the best performances in total force compared with traditional markerless design with clear skin and marker design with dense ink design. The enhanced force prediction performance is verified in both single image prediction (ResNet) and sequential image prediction (ResNet+ConvGRU). Specifically, force prediction errors are improved 66\% using grey squares compared to dense ink design in unseen objects with sequential images and 36\% compared with clear image. In the seen groups, the improvement is also around 56\% using a grey square with sequential imaged compared to traditional design. In addition, we also show force prediction results for 500 images with our two translucent design in Fig. \ref{fig_force_500}, both demonstrating small force errors in three axes. The above results validate the superiority of our MagicSkin in the prediction of 3-axis forces.

\begin{table}[]
\centering
\caption{Force Prediction Performance in Seen Objects}
\label{tab:seen}
\resizebox{0.48\textwidth}{!}{%
\begin{tabular}{|c|c|cc|cc|cc|c|}
\hline
\multirow{2}{*}{Method}                                                      & \multirow{2}{*}{\begin{tabular}[c]{@{}c@{}}Elastomer \\ Type\end{tabular}} & \multicolumn{2}{c|}{$F_x$}                                & \multicolumn{2}{c|}{$F_y$}                                & \multicolumn{2}{c|}{$F_z$}                                & Total Force          \\ \cline{3-9} 
                                                                             &                                                                            & \multicolumn{1}{c|}{$error(N)\downarrow$} & $R^2\uparrow$ & \multicolumn{1}{c|}{$error(N)\downarrow$} & $R^2\uparrow$ & \multicolumn{1}{c|}{$error(N)\downarrow$} & $R^2\uparrow$ & $error(N)\downarrow$ \\ \hline
\multirow{4}{*}{ResNet}                                                      & Clear (markerless)                                                         & \multicolumn{1}{c|}{0.2421}               & 0.36          & \multicolumn{1}{c|}{0.2513}               & 0.48          & \multicolumn{1}{c|}{\textbf{0.5823}}      & 0.84          & 0.5560               \\
                                                                             & Dense ink (marker)                                                         & \multicolumn{1}{c|}{\textbf{0.1203}}      & \textbf{0.89} & \multicolumn{1}{c|}{\textbf{0.1486}}      & \textbf{0.88} & \multicolumn{1}{c|}{0.6358}               & \textbf{0.86} & 0.5884               \\
                                                                             & Grey lines (translucent)                                                   & \multicolumn{1}{c|}{0.2132}               & 0.66          & \multicolumn{1}{c|}{0.1532}               & 0.84          & \multicolumn{1}{c|}{1.1741}               & 0.73          & 1.0645               \\
                                                                             & Grey squares (translucent)                                                 & \multicolumn{1}{c|}{0.1528}               & 0.66          & \multicolumn{1}{c|}{0.2482}               & -0.22         & \multicolumn{1}{c|}{0.5887}               & 0.73          & \textbf{0.5130}      \\ \hline
\multirow{4}{*}{\begin{tabular}[c]{@{}c@{}}ResNet\\ +\\ ConvGRU\end{tabular}} & Clear (markerless)                                                         & \multicolumn{1}{c|}{0.2803}               & 0.05          & \multicolumn{1}{c|}{0.3158}               & 0.01          & \multicolumn{1}{c|}{0.6215}               & 0.81          & 0.5483               \\
                                                                             & Dense ink (marker)                                                         & \multicolumn{1}{c|}{\textbf{0.1058}}      & \textbf{0.93} & \multicolumn{1}{c|}{\textbf{0.1122}}      & \textbf{0.94} & \multicolumn{1}{c|}{0.6285}               & 0.90          & 0.5690               \\
                                                                             & Grey lines (translucent)                                                   & \multicolumn{1}{c|}{\textbf{0.0895}}      & \textbf{0.93} & \multicolumn{1}{c|}{0.1125}               & 0.90          & \multicolumn{1}{c|}{\textbf{0.4051}}      & 0.91          & 0.3597               \\
                                                                             & Grey squares (translucent)                                                 & \multicolumn{1}{c|}{0.0901}               & 0.89          & \multicolumn{1}{c|}{\textbf{0.0884}}      & 0.92          & \multicolumn{1}{c|}{\textbf{0.2708}}      & \textbf{0.93} & \textbf{0.2475}      \\ \hline
\end{tabular}%
}
\end{table}

\begin{table}[]
\centering
\caption{Force Prediction Performance in Unseen Objects}
\label{tab:unseen}
\resizebox{0.48\textwidth}{!}{%
\begin{tabular}{|c|c|cc|cc|cc|c|}
\hline
\multirow{2}{*}{Method}                                                      & \multirow{2}{*}{\begin{tabular}[c]{@{}c@{}}Elastomer \\ Type\end{tabular}} & \multicolumn{2}{c|}{$F_x$}                                & \multicolumn{2}{c|}{$F_y$}                                & \multicolumn{2}{c|}{$F_z$}                                & Total Force          \\ \cline{3-9} 
                                                                             &                                                                            & \multicolumn{1}{c|}{$error(N)\downarrow$} & $R^2\uparrow$ & \multicolumn{1}{c|}{$error(N)\downarrow$} & $R^2\uparrow$ & \multicolumn{1}{c|}{$error(N)\downarrow$} & $R^2\uparrow$ & $error(N)\downarrow$ \\ \hline
\multirow{4}{*}{ResNet}                                                      & Clear (markerless)                                                         & \multicolumn{1}{c|}{0.2609}               & 0.18          & \multicolumn{1}{c|}{0.3582}               & -0.34         & \multicolumn{1}{c|}{0.7456}               & 0.69          & 0.7317               \\
                                                                             & Dense ink (marker)                                                         & \multicolumn{1}{c|}{0.4734}               & -0.54         & \multicolumn{1}{c|}{0.3493}               & \textbf{0.39} & \multicolumn{1}{c|}{1.1487}               & 0.52          & 1.1050               \\
                                                                             & Grey lines (translucent)                                                   & \multicolumn{1}{c|}{0.3741}               & -0.11         & \multicolumn{1}{c|}{0.3807}               & -1.21         & \multicolumn{1}{c|}{1.0853}               & 0.41          & 1.0012               \\
                                                                             & Grey squares (translucent)                                                 & \multicolumn{1}{c|}{\textbf{0.2430}}      & \textbf{0.23} & \multicolumn{1}{c|}{\textbf{0.3083}}      & 0.01          & \multicolumn{1}{c|}{\textbf{0.8003}}      & \textbf{0.71} & \textbf{0.7296}      \\ \hline
\multirow{4}{*}{\begin{tabular}[c]{@{}c@{}}ResNet\\ +\\ ConvGRU\end{tabular}} & Clear (markerless)                                                         & \multicolumn{1}{c|}{0.2588}               & 0.05          & \multicolumn{1}{c|}{0.2924}               & 0.00          & \multicolumn{1}{c|}{0.8408}               & 0.71          & 0.7926               \\
                                                                             & Dense ink (marker)                                                         & \multicolumn{1}{c|}{\textbf{0.1981}}      & \textbf{0.76} & \multicolumn{1}{c|}{\textbf{0.1429}}      & \textbf{0.89} & \multicolumn{1}{c|}{1.6325}               & 0.40          & 1.5693               \\
                                                                             & Grey lines (translucent)                                                   & \multicolumn{1}{c|}{0.2113}               & 0.38          & \multicolumn{1}{c|}{0.1407}               & 0.77          & \multicolumn{1}{c|}{\textbf{0.6746}}      & 0.71          & 0.6579               \\
                                                                             & Grey squares (translucent)                                                 & \multicolumn{1}{c|}{0.1371}               & 0.74          & \multicolumn{1}{c|}{0.1424}               & 0.83          & \multicolumn{1}{c|}{\textbf{0.5416}}      & \textbf{0.87} & \textbf{0.5012}      \\ \hline
\end{tabular}%
}
\end{table}

\section{Discussion and Conclusion}\label{Sec:DiscussionConclusion}

\subsection{Discussion}\label{Discussion}

The experiments provide strong evidence that tinted, translucent markers address the long-standing trade-off between tangential displacement tracking and preservation of surface detail in vision-based tactile sensing. Conventional dense ink markers, although simple to fabricate, obscure fine geometric features of the contact surface and consequently degrade downstream performance. This drawback is reflected in both reduced texture classification accuracy and unstable optical flow retention. Grey line patterns, designed to improve retention by providing high-contrast edges, offered initial advantages but deteriorated under repeated use and high strain, highlighting susceptibility to wear. In contrast, the grey square design consistently maintained retention above 90\% at strains exceeding 15\% while preserving a low forward-backward (FB) error, indicating robustness to both deformation and prolonged operation.

\begin{figure}[htbp]
\centering
\includegraphics[width=0.46\textwidth]{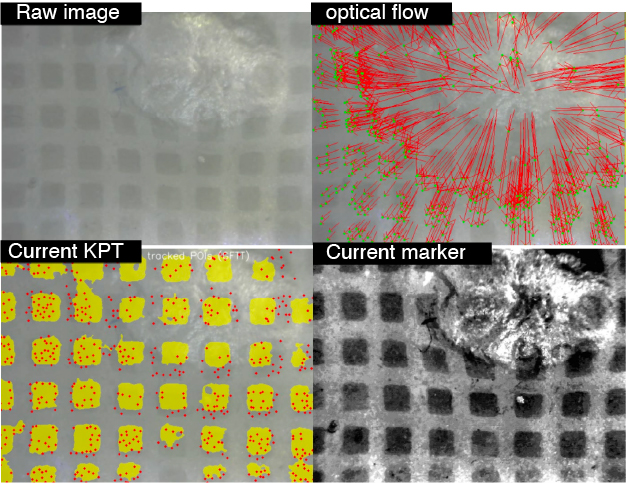}
\caption{Tangential displacement tracking in grey square (translucent) remains successful under large deformation with a strawberry contact. Despite the markers within the contact area becoming visually indistinct, optical flow features remain stable.}
\label{fig_strawberry_marker}
\end{figure}

A closer analysis of optical flow behaviour clarifies the source of this improvement. Translucent markers embedded within the elastomer exhibit lower contrast with the background than dense ink patterns, producing masks that are less sharply defined. Although this might appear disadvantageous, the reduced contrast allows the detection of a larger number of distinct, trackable features within the same region. These features are reliably tracked across frames, yielding both higher point counts and improved retention. Dense ink, in contrast, yields crisp, high-contrast boundaries that can locally minimise FB error, but provide few unique features for long-term tracking, limiting stability. As a result, translucent markers achieve a more favourable balance: substantially higher retention than dense ink while maintaining comparable FB error, providing a more reliable basis for lateral displacement estimation. This behaviour is exemplified in Fig.~\ref{fig_strawberry_marker}, where optical flow remains stable under large strains even when markers within the contact region become visually indistinct; the figure also illustrates the processing pipeline, including the extracted mask, keypoint selection within each mask, and the contrast-enhanced frame used for Lucas-Kanade tracking. Further gains are feasible by introducing simple gating to reject keypoints with FB error above a fixed threshold; because the translucent design yields a larger pool of candidate features, such gating would not materially reduce measurement accuracy.

The advantages of translucent markers extend beyond optical flow tracking to classification and force estimation. Object recognition accuracy improved from 98.3\% with dense ink to 99. 2\% with grey squares, while texture classification increased from 63.9\% to 93.5\%. The multiaxis force prediction error was reduced by more than 66\% with consistently high coefficients of determination ($R^2$), underscoring the value of preserving surface deformation signals along with robust tracking. Collectively, these results demonstrate that translucent marker designs enhance robustness, support more discriminative feature learning, and improve force estimation accuracy, establishing a practical and scalable approach for vision-based tactile sensing.

\begin{figure}[htbp]
\centering
\includegraphics[width=0.45\textwidth]{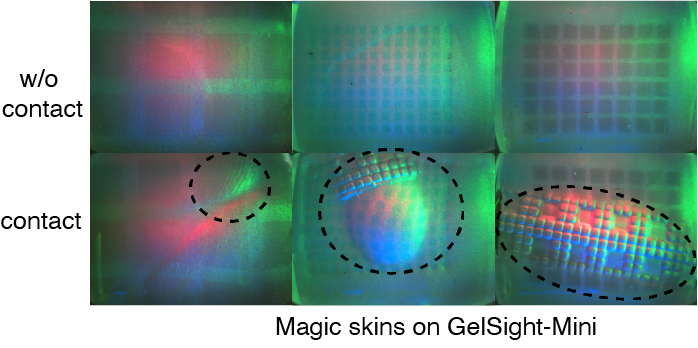}
\caption{MagicSkin as a plug-in module for GelSight systems such as GelSight-mini.}
\label{fig_texture}
\end{figure}

In summary, the findings validate the central hypothesis of this study: tinted, translucent markers eliminate the inherent trade-off between shear-force tracking and deformation visibility in VBTS. By enabling stable lateral displacement tracking without sacrificing surface detail, the proposed design overcomes the limitations of prior marker-based and markerless approaches, establishing a practical and generalizable path forward for next-generation tactile sensors.

\subsection{Future Work}\label{FutureWork}

Future work will evaluate the sensor with a broader set of objects, including non-rigid bodies, to assess performance under deformable contact geometries. A medium-term direction is the development of image processing pipelines for tinted, translucent elastomers under multicolour illumination, where the marker hue varies with lighting (Fig.~\ref{fig_texture}). Reliable segmentation in these conditions would enable MagicSkin to support photometric stereo for combined geometry–force reconstruction, and integration with NormalFlow for 6DoF pose tracking \cite{Huang2024}, advancing applications such as slip detection and contact-rich manipulation. Further refinement of marker designs will also be explored to minimise transparency artefacts under high strain, improving robustness. The fabrication method can also be extended to generate smaller, denser, or randomised marker patterns, increasing spatial coverage while retaining surface texture information to support optical flow tracking and high-resolution contact reconstruction. In the longer term, these developments will facilitate embedding the sensor within closed-loop controllers and multi-fingered robotic grippers to achieve dexterous in-hand manipulation in unstructured environments.

\subsection{Conclusion}\label{Conclusion}

This work introduced tinted, translucent markers that resolve the trade-off between tangential displacement tracking and deformation visibility in VBTS. The grey square design delivered the strongest performance, sustaining more than 90\% point retention under high strain with sub-millimetre accuracy and the lowest mean forward–backward error among all skins. Classification also improved, with object accuracy above 99\% and texture recognition reaching 93.5\%. The force prediction error decreased by more than 66\% relative to dense ink while maintaining a high $R^2$ across the axes. These results show that translucent markers improve tracking stability, preserve surface detail for recognition, and improve force estimation within a single, simple skin that requires no additional hardware or software. The findings confirm the central hypothesis and indicate a practical path toward reliable slip detection, in-hand manipulation, and multi-contact robotic control.
    
\bibliographystyle{IEEEtran}  
\bibliography{manuscript}

\end{document}